\documentclass[conference]{IEEEtran}
\IEEEoverridecommandlockouts

\usepackage{cite}
\usepackage{amsmath,amssymb,amsfonts}
\usepackage{algorithmic}
\usepackage{graphicx}
\usepackage{textcomp}
\usepackage[dvipsnames,svgnames,x11names]{xcolor}
\def\BibTeX{{\rm B\kern-.05em{\sc i\kern-.025em b}\kern-.08em
    T\kern-.1667em\lower.7ex\hbox{E}\kern-.125emX}}

\usepackage{ulem}
\usepackage{comment}
\usepackage{tcolorbox}
\usepackage{color}
\usepackage{multirow}

\begin{document}

\title{ Tracing Partisan Bias to Its Emotional Fingerprints: A Computational Approach to Mitigation \\

\thanks{\dag \ These authors contributed equally to this work.

This work is partially supported by European Research Council (ERC) under the European Union's Horizon 2020 research and innovation programme (grant agreement No. 101002240); Guangdong Provincial Key Laboratory of Interdisciplinary Research and Application for Data Science, Beijing Normal Hong Kong Baptist University., project code 2022B1212010006, UIC research grant R0400001-22; National Natural Science Foundation of China (No.12231004), UIC research grant UICR0600048; National Natural Science Foundation of China (No.1272054), UIC research grant UICR0600036; Guangdong University Innovation and Enhancement Programme Funds Featured Innovation Project 2018KTSCX278. The computations in this paper were performed using the Bayes Cluster (USBC) provided by the Department of Statistics and Data Science, Beijing Normal Hong Kong Baptist University. We also thank Kylan Rutherford for helpful comments on the manuscript.}
}

\author{\IEEEauthorblockN{Junjie Liu\textsuperscript{\dag, a}, Xi Luo\textsuperscript{\dag, b,c,d}, Sirong Wu\textsuperscript{b,c,d}, Gengchen Sun, \textsuperscript{b,c,d, e}, Yuhui Deng\textsuperscript{b,c}}

\IEEEauthorblockA{\textsuperscript{a}\textit{Department of Political Science, Trinity College Dublin, Dublin, Ireland}}
\IEEEauthorblockA{\textsuperscript{b}\textit{Guangdong Provincial Key Laboratory of Interdisciplinary Research and Application for Data Science, Zhuhai, China}} 
\IEEEauthorblockA{\textsuperscript{c}\textit{Department of Statistics and Data Science, Beijing Normal Hong Kong Baptist University., Zhuhai, China}} 
\IEEEauthorblockA{\textsuperscript{d}\textit{Faculty of Science, Hong Kong Baptist University, Hong Kong SAR, China}}
\IEEEauthorblockA{\textsuperscript{e}\textit{Hong Kong aiKnow Limited, Hong Kong SAR, China}}
\IEEEauthorblockA{\{liuj13\}@tcd.ie, \{xiluo, sirongwu, sungengchen, ivandeng\}@uic.edu.cn}

}

\maketitle
\begin{figure}
    \centering
    \includegraphics[width=0.7\linewidth]{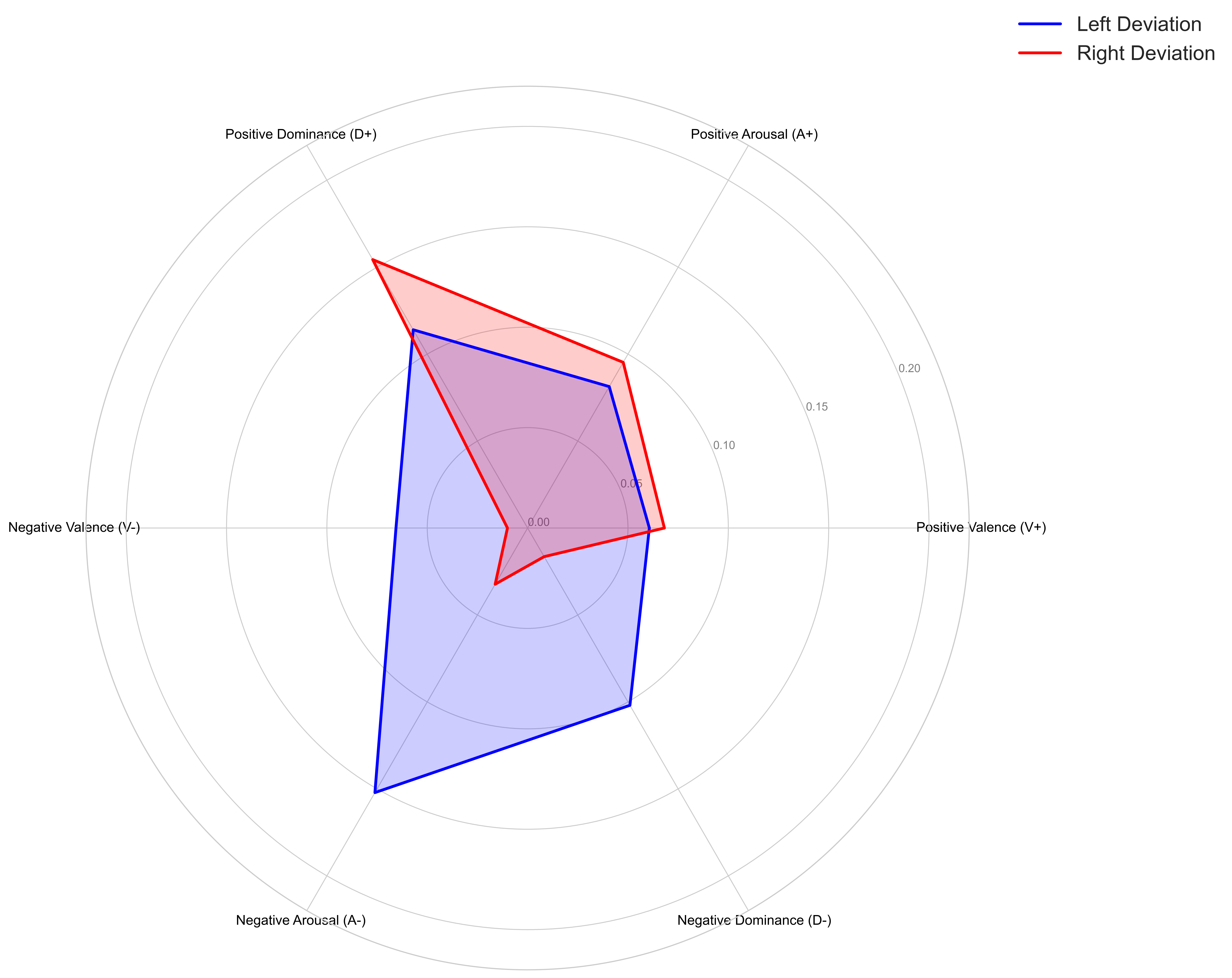}
    \caption{Emotional and Lexical Deviation from the Neutral Baseline. The chart visualizes the distinct emotional fingerprints of \textcolor{blue}{Left-leaning} and \textcolor{red}{Right-leaning} media as deviations from the centre. The opposing shapes reveal their symmetrical but contrary emotional amplification strategies.}
    \label{fig:difference_radar_final}
\end{figure}

\begin{abstract}
This study introduces a novel framework for analyzing and mitigating media bias by tracing partisan stances to their linguistic roots in emotional language. We posit that partisan bias is not merely an abstract stance but materializes as quantifiable 'emotional fingerprints' within news texts. These fingerprints are systematically measured using the Valence-Arousal-Dominance (VAD) framework, allowing us to decode the affective strategies behind partisan framing. Our analysis of the Allsides dataset confirms this hypothesis, revealing distinct and statistically significant emotional fingerprints for left, centre, and right-leaning media. Based on this evidence-driven approach, we then propose a computational approach to mitigation through NeutraSum, a model designed to neutralise these identified emotional patterns. By explicitly targeting the VAD characteristics of biased language, NeutraSum generates summaries that are not only coherent but also demonstrably closer to an emotionally neutral baseline. Experimental results validate our framework: NeutraSum successfully erases the partisan emotional fingerprints from its summaries, achieving a demonstrably lower emotional bias score than other models. This work pioneers a new path for bias mitigation, shifting the focus from treating symptoms (political labels) to addressing the cause: the emotional encoding of partisan bias in language.
\\
\textcolor{red}{Declaimer: This study involves the generation of synthetic news content using with language model. The generated content was created solely for the purposes of this research and does not represent actual news or the views of any individual, organisation, or entity.} 
\end{abstract}


\section{Introduction}
The digital revolution has unleashed an unprecedented flood of information, posing a significant challenge for media consumers to navigate and evaluate content effectively. The exponential growth of online news platforms and social media channels has made it increasingly difficult to discern relevant and reliable information. This deluge of content not only overwhelms readers but also amplifies implicit polarisation, as audiences gravitate toward material that aligns with their pre-existing beliefs—whether by choice or through algorithmic personalisation~\cite{eppler:15, schmitt:18}. Such selective exposure fosters ideological echo chambers, eroding the diversity of perspectives essential for informed decision-making. Media bias further intensifies the problem, as journalists often frame stories to support specific political agendas, leading to starkly different portrayals of the same events~\cite{DA:20, Rodrigo:24, Gentzkow:06}. This selective framing encourages readers to engage with content that reinforces their political inclinations, perpetuating stereotypes and distorting critical decisions~\cite{Sunstein:01}. As a result, finding effective strategies to mitigate media bias in news reporting has become a pressing research priority.

To address this issue, the concept of a balanced news diet~\cite{van2021fighting, muise2022quantifying, angelucci2024media} has gained prominence, advocating for the consumption of diverse viewpoints to foster a more comprehensive understanding of events. However, achieving this balance is non-trivial, as it requires not only access to politically diverse sources but also the ability to distill and synthesis information effectively. A feasible systemic research framework to mitigate media bias from news is news aggregation~\cite{Park:11, Lee:22, Zhang:19, Bang:23}. It provides a comprehensive perspective of an event to mitigate the exposure of the far left or far right slanted bias. Based on this approach, one early study~\cite{Lee:22} proposed a new task to generate a neutral summary from a triplet of news articles reporting on the same issues. To achieve the goal of news neutralisation, they utilised a hierarchical framework from title to articles to generate a bias-free summary. A further refined work~\cite{Bang:23} introduced polarity loss to minimize the polarity difference in the input articles. Although these studies contribute valuable insights, they also exhibit notable limitations. Although indicative information and polarity regularisation techniques were explored to produce bias-controlled summaries, their effectiveness in achieving genuine neutrality remains uncertain. For instance, while title neutralisation can serve as a helpful subtask, it does not directly address the inherent bias embedded in the summarisation process. Moreover, minimizing polarity differences poses significant challenges, as even news articles covering the same event often present different factual elements, making strict mappings across sources difficult to establish.

We argue that partisan bias is not merely a matter of political stance but manifests concretely through the use of emotional language. In particular, we posit that news outlets encode bias in the form of quantifiable \emph{emotional fingerprints}, which can be systematically decoded using the Valence–Arousal–Dominance (VAD) framework~\cite{picard1997ective, Moham:18}. Unlike simple polarity, VAD captures the multidimensional affective strategies underlying partisan framing—such as inciting anger (low valence, high arousal) or projecting authority (high dominance). This perspective reframes the task of bias mitigation as one of neutralising measurable emotional patterns, rather than merely balancing ideological polarity. Guided by this insight, we propose \textbf{NeutraSum}, a novel framework for multi-document summarisation that explicitly targets emotional fingerprints. NeutraSum augments a pre-trained BART backbone with two task-specific neutrality losses: the \emph{Equal-Distance Loss}, which enforces symmetry by positioning neutral summaries equidistant from left- and right-leaning embeddings; and the \emph{Contrastive Loss}, which helps align generated summaries with expert-written references while distancing them from polarised inputs. By combining these neutrality constraints with standard multi-document summarisation objectives, NeutraSum generates outputs that preserve salient content while remaining demonstrably closer to an emotionally neutral baseline.


To better measure media bias, we explore attitudes towards specific topics or figures as a proxy to measure political bias in media reports, since some media outlets, in their reporting, convey their political stance by either criticising or endorsing the speech, actions of target political figures or the social movements~\cite{Elejalde:18}. We use chain-of-thought~\cite{Wei:22} as an LLM-based evaluator to identify attitude-reflecting words, and a lexicon-based rating corpus to assess their polarity. More specifically,  Chain-of-Thought (CoT) could be utilised to answer a series of sub-questions through intermediate reasoning steps. Through these sub-answers, LLM could understand the context and implicit thoughts towards the events, which is helpful to identify the relevant words that reflect attitudes toward a particular topic or character in the media reports and quantify the selected words with the VAD corpus~\cite{Moham:18}.

Our experiments are conducted on the Allsides dataset~\cite{Lee:22}, which incorporates same-story news articles from different political leanings, as well as expert-written summaries. Fine-tuning the neutrality losses significantly guides the generative model toward producing more neutral outputs. The results indicate that these losses effectively reduce media bias while consistently preserving the salient information in the summaries. This research is to pioneer a new path for bias mitigation: instead of addressing abstract political labels, we tackle partisan bias at its linguistic roots by identifying, quantifying, and actively neutralising the ``emotional fingerprints" embedded within news texts.

\section{Related Work}

\subsection{Media Bias Detection}

News journalists often frame their news stories to omit or exaggerate some parts of the facts to support their political leaning, a phenomenon commonly referred to as media bias~\cite{Hamborg:19}. A variety of computational approaches have been developed to identify media bias by extracting the polarity features~\cite{Liu:22a, Baly:20}, which assists in predicting the political leaning. It has also been found that headlines could act as additional cues to classify media bias~\cite{Gangula:19}. Besides identifying the polarity of the news articles, some more refined datasets~\cite{Spinde:21, Fan:19} annotated at the sentence and lexicon level are leveraged by recent research. The study of media bias is increasingly intertwined with the analysis of intrinsic biases within Large Language Models (LLMs) themselves, as these models are trained on vast corpora of media texts and inevitably inherit their biases. Recent work such as Peng et al.~\cite{peng2024beyond} has moved beyond simple left-right classifications, proposing multi-dimensional frameworks to evaluate the political behavior of LLMs. They found that most models exhibit a centre-left or left-leaning ideological stance in their content and topic engagement. This macro-level observation raises a critical question that our work seek to address: how does this documented ideological leaning manifest at the micro-level of linguistic style? While Peng et al. analyse what LLMs say, our research focuses on how they say it.

While these polarity-centred methods and recent LLM-based advances have contributed significantly to bias detection, they remain inherently limited in capturing the nuanced emotional strategies employed in partisan framing. For instance, two articles can be equally negative in polarity, yet one may aim to incite anger (high arousal), while the other evokes sadness (low arousal). To address this, our work moves beyond polarity and leverages the multidimensional Valence-Arousal-Dominance (VAD) framework~\cite{Moham:18}. While VAD has been applied in affective computing, its use in systematically deconstructing the ‘emotional fingerprints’ of partisan bias in news media remains largely unexplored. A notable exception is the work of Lee et al.~\cite{Lee:22}, who pioneered the use of Arousal to measure sensationalism. Our research builds upon and significantly extends this direction by utilising the full three-dimensional VAD space to trace partisan stances to their complex emotional signatures.

\subsection{Media Bias Mitigation}
Despite the current absence of large-scale media annotation datasets including bias span and expert-written summaries,  alone with standardised measurements for media bias mitigation, numerous studies have still endeavoured to reduce media bias in news coverage~\cite{Park:09, Park:11, Trampuvs:15, Lee:21, Lee:22}. A GAN framework~\cite{Liu:21a} was designed to reverse or neutralise the political polarity of the news articles. However, the definition of neutrality adopted is limited to the ``centre'' media rating. It is insufficient to assert a reduction in media bias, as a centre stance does not inherently equal bias-free~\cite{Allsides:04}. They further deployed a depolarisation algorithm~\cite{Liu:21b} to mitigate bias in polarised texts, while the measurement of media bias was conducted using both manual and model-based methods, which could not be considered as a universe of media bias metrics since no consistently reliable predicted results could be produced. Another emerging mitigation task generates a neutral summary based on the same-story polarity news articles~\cite{Lee:21, Lee:22, Bang:23}. However, they did not provide clear neutralisation guidance or a mechanism to reduce media bias.

\section{VAD Scale shows the differences}
The VAD analysis on the Allsides dataset, summarised in Table~\ref{tab:vad_fingerprints}, reveals that partisan bias is not evenly distributed across the emotional spectrum but is asymmetrically concentrated in the expression of \emph{negative emotion}. The Analysis of Variance (ANOVA) confirms this divide: while no significant differences were observed in positive emotional metrics ($p > 0.05$), we found strong effects for Negative Valence ($p=0.011$), Negative Arousal ($p=0.007$), and Negative Dominance ($p=0.010$). To isolate the main driver, we conducted a Tukey HSD post-hoc test on the overall Arousal Score ($p=0.042$). The results were unequivocal: left-leaning media exhibited a significantly higher mean arousal compared to the centre baseline ($p=0.032$), while other pairwise comparisons were not significant. This provides robust, quantifiable evidence that the most distinctive partisan fingerprint is the \emph{``Forceful Critique''} strategy of the Left, characterised by sustained use of high-intensity, negatively valence language. Such a channel becomes a critical target for any effective mitigation strategy.

Importantly, this micro-level linguistic evidence complements macro-level observations of ideological bias in language models. For example, Peng~\cite{peng2024beyond} documents that many LLMs trained on news corpora exhibit a centre-left orientation. Our findings suggest a potential mechanism: the left-leaning ideological content often identified in models may be underpinned by a statistically verifiable rhetorical style—the systematic amplification of negative arousal. In other words, partisan leaning is not merely topical, but encoded and expressed through distinct emotional fingerprints.  This insight reframes the design of debiasing methods. Rather than treating neutrality as the absence of polarity labels, we operationalise it as the suppression of these measurable emotional fingerprints. In Section \ref{sec:method}, we translate this principle into two neutrality-driven loss functions—Equal-Distance and Contrastive Loss—that guide our proposed NeutraSum framework.

\begin{table}[h!]
\centering
\begin{tabular}{l|ccc} \hline
\textbf{VAD Metric} & \textbf{Left} & \textbf{centre} & \textbf{Right} \\ \hline
\multicolumn{4}{c}{\textit{Overall Emotional Scores}} \\
V\_Score & \textbf{4.94} & 4.81 & 4.89 \\
A\_Score & \textbf{7.52} & 7.29 & 7.42 \\
D\_Score & \textbf{8.85} & 8.63 & 8.80 \\ \hline
\multicolumn{4}{c}{\textit{Positive Emotional Components}} \\
V\_POSITIVE & 3.52 & 3.46 & \textbf{3.53} \\
A\_POSITIVE & 4.96 & 4.88 & \textbf{4.97} \\
D\_POSITIVE & 7.03 & 6.92 & \textbf{7.07} \\ \hline
\multicolumn{4}{c}{\textit{Negative Emotional Components}} \\
V\_NEGATIVE & \textbf{1.42} & 1.35 & 1.36 \\
A\_NEGATIVE & \textbf{2.56} & 2.41 & 2.44 \\
D\_NEGATIVE & \textbf{1.82} & 1.71 & 1.73 \\ \hline
\end{tabular}
\caption{Mean Emotional Fingerprints Across Political Leanings. The table displays the average VAD scores for each media category. Bold values highlight the highest score for each metric among the partisan groups (Left and Right), demonstrating their respective emotional amplification strategies relative to the centre baseline.}
\label{tab:vad_fingerprints}
\end{table}

\begin{figure*}[h!]
    \centering
    \includegraphics[width=0.6\linewidth]{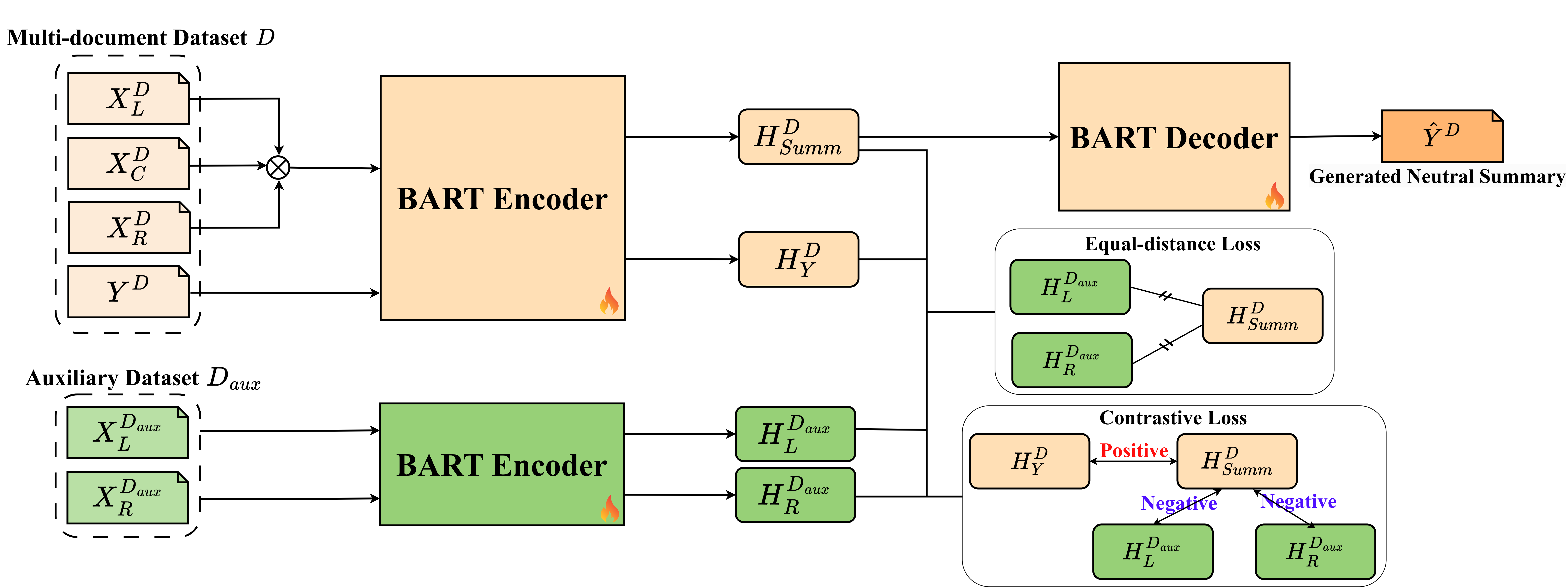}
    \caption{Illustration of the training process of our model NeutraSum. The main task is to summarise a neutral summary $\hat{Y}^{D}$ from a Multi-document dataset $D$ using BART pre-trained language model. This dataset incorporates triplets of left, centre, and right-wing news articles $\left\{ X^{D}_{L}, X^{D}_{C}, X^{D}_{R} \right\}$, as well as expert-written summaries $Y^{D}$. Each triplet is reporting on the same issue. In the encoder framework, we introduced an Auxiliary Dataset $D_{\text{aux}}$ to propose polarised left-wing and right-wing news articles $X^{D_{\text{aux}}}_{L}$, $X^{D_{\text{aux}}}_{R}$. This dataset aims to join two neutrality losses (Equal-distance Loss and Contrastive Loss) by finetuning the whole semantic space. These two datasets jointly guide the neutral writing of a summary.}
    \label{fig:media-bias-model}
\end{figure*}

\vspace{-0.7cm}

\section{Methodology}\label{sec:method}
In this section, we introduce the details of our \textbf{NeutraSum} model. To achieve good summarisation performance while minimizing lower media bias, our model is governed by a multi-document summarisation loss to generate high-quality summaries. Additionally, two neutrality losses (contrastive loss and equal-distance loss) are employed to mitigate media bias in summaries. The model's architecture is shown in Fig. \ref{fig:media-bias-model}.

\subsection{Multi-document summarisation Loss ($L_{\text{MDS}}$)}\label{AA}
To align with the neutral summarisation task, MDS-Loss is the main loss to ensure coherent and content-focused writing by conditional generation. To ensure the comprehensiveness and accuracy of the news, multi-document summarisation would be utilised to aggregate left, centre, and right-wing media coverage to generate a de-bias summary. In the training phase, the generation process would learn the neutral writing features by following the texts of the expert-written summary $Y^{D}$. The MDS-Loss could be illustrated as follows:

{\small
\begin{align}
    X^{D} &= Concat(X_{L}^{D}, X_{C}^{D}, X_{R}^{D}) = \{x_{i}^{D}\}_{i=1}^{|T|} \\
    Y^{D} &= \{y_{i}^{D}\}_{i=1}^{|Y|} \\
    H^{D}_{\text{summ}} &= \text{Enc}_{\text{summ}}(X^{D}) \\
    O^{D}_{\text{summ}} &= \text{Dec}_{\text{summ}}(H^{D}_{\text{summ}}) \\
    \hat{Y}^{D} &= W_{H}O^{D}_{\text{summ}} + b_{H} \\
    L_{MDS} &= CE(\hat{Y}^{D}, Y^{D}) = -\sum_{i=1}^{|Y|}{y_{i}^{D}log\hat{y}_{i}^{D}}
\end{align}
}

$X_{L}^{D}$, $X_{C}^{D}$, $X_{R}^{D}$ represents the left-, centre-, and right-wing news articles relevant to the same event. $D$ is denoted as the Allsides dataset. $x_i^{D}$ is the concatenation of the tokenised articles, where $X^{D} = \{x_i^{D}\}_{i=1}^{|T|}$ with the length is $|T|$. The $Y^{D}$ represents the corresponding expert-written summary and tokenised into $\{y_{i}^{D}\}_{i=1}^{Y}$ with the length of $|Y|$.

$\text{Enc}_{\text{summ}}$ and $\text{Dec}_{\text{summ}}$ demonstrate the encoder and decoder part of the summarisation BART architecture. Input articles $X^{D}$ would go through the encoder framework $\text{Enc}_{\text{summ}}$ to obtain the hidden state $H^{D}_{\text{summ}} \in R^{|T| \times d}$ in a $d$-dimensional semantic embedding, which is default to 1024. Then $H^{D}_{\text{summ}}$ would feed into the decoder framework  $\text{Dec}_{\text{summ}}$ to get the hidden state $O^{D}_{\text{summ}} \in R^{|Y| \times d}$. Finally, to obtain the probabilities of each generated token, the decoder hidden state $O^{D}_{\text{summ}}$ would be projected to a $w$-dimensional semantic embedding  $\hat{Y}^{D} \in R^{|Y| \times w}$ through the weight $W_{H}$ and bias $b_{H}$, it could also be considered as a predicted probability distribution for the predicted tokens. Since the predicted output should follow the expert-written summary $Y^{D}$, the multi-document summarisation loss $L_{\text{MDS}}$ would be calculated by the cross-entropy loss $CE$ between expert-written summary $Y^{D}$ and projected output $\hat{Y}^{D}$ to optimise the model.

\subsection{Equal-Distance Loss ($L_{\text{ED}}$)}

To neutralise a summary, it must inherit neither of left-leaning nor right-leaning emotional signatures. Therefore, we introduce the Equal-Distance Loss $L_{\text{ED}}$ This loss function operationalizes the principle of emotional neutrality in the semantic space. It posits that an emotionally neutral summary embedding $\bar{H}_{\text{summ}}^{D}$ should lie equidistant from the semantic embedding of left-side $\bar{H}_{\text{L}}^{D_{\text{aux}}}$ and right-side $\bar{H}_{\text{R}}^{D_{\text{aux}}}$ articles from the auxiliary dataset $D_{\text{aux}}$. Specifically, the loss is defined as:

\begin{footnotesize}
\begin{align}
    &H_{L}^{D_{\text{aux}}}, H_{R}^{D_{\text{aux}}} = Enc_{\text{aux}}(X_{L}^{D_{\text{aux}}}), Enc_{\text{aux}}(X_{R}^{D_{\text{aux}}}) \\
    &[\bar{H}_{L}^{D_{\text{aux}}}, \bar{H}_{R}^{D_{\text{aux}}}, \bar{H}_{\text{summ}}^{D}] = Avg([H_{L}^{D_{\text{aux}}}, H_{R}^{D_{\text{aux}}}, H_{\text{summ}}^{D}])\\
    &L_{\text{ED}} = |sim(\bar{H}_{L}^{D_{\text{aux}}}, \bar{H}_{\text{summ}}^{D}) - sim(\bar{H}_{R}^{D_{\text{aux}}},\bar{H}_{\text{summ}}^{D})|_{1} \label{eq:equal_dist_loss}
\end{align}
\end{footnotesize}

where $sim(\cdot, \cdot)$ denotes the cosine similarity, and the $L_1$ norm ensures the loss is positive. The auxiliary encoder model $Enc_{\text{aux}}$ which adopts the BART encoder module, is designed to align the semantic space of left-side and right-side news articles, denoted as $X_{L}^{D_{\text{aux}}}$ and $X_{R}^{D_{\text{aux}}}$, are fed into the auxiliary encoder to obtain their respective embedded hidden states $H_{L}^{D_{\text{aux}}}$ and $X_{R}^{D_{\text{aux}}}$. To capture the general semantic meaning of these articles, the embedded hidden states are averaged across the batch dimension, resulting in $\bar{H}_{L}^{D_{\text{aux}}}$ and $\bar{X}_{R}^{D_{\text{aux}}}$. Similarly, the hidden states of the summarisation task are average to produce the summarisation embedding $H_{summ}^{D}$. By minimizing $L_{\text{ED}}$, the optimisation ensure that $\bar{H}_{\text{summ}}$ lies on the bisector hyperplane, which geometrically separates $\bar{H}_{\text{L}}^{D_{\text{aux}}}$ and $\bar{H}_{\text{R}}^{D_{\text{aux}}}$, enforcing symmetry. From a theoretical perspective, the $L_{\text{ED}}$  enforces Lipschitz continuity, as cosine similarity is a smooth function bounded between $[-1, 1]$, and the $L_1$ regularisation does not introduce abrupt changes. This guarantees that small perturbations in $\bar{H}_{\text{summ}}$ result in small changes in the loss value, stabilizing the optimisation process.

\begin{figure*}[!t]
    \centering
    \includegraphics[width=1.0\linewidth]{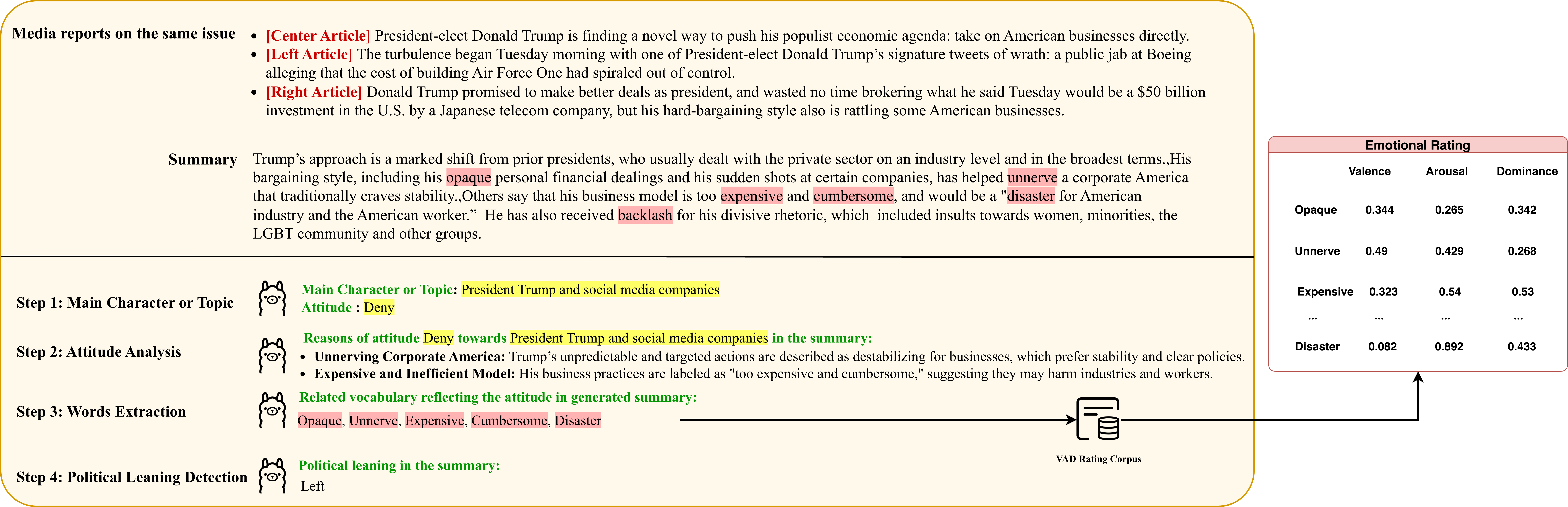}
    \caption{An illustration of our Chain-of-Thought LLM-based Metric framework. Given media reports on the same issue and a generated summary, the framework employs a step-by-step reasoning process to assess media bias. It first identifies the main characters or topics and their associated attitudes, then analyses the underlying rationale behind these perspectives. Next, it extracts key words or phrases that reflect the detected attitudes and, ultimately, determines the political leaning of the text. After the chain-of-thought reasoning process, the framework leverages the VAD rating corpus to quantify the emotional tone of the extracted attitude-indicative words (from step 3) based on their Valence, Arousal, and Dominance scores.}
    \label{fig:cot-framework}
\end{figure*}

\subsection{Contrastive Loss ($L_{\text{Con}}$)}

While the Equal-Distance Loss prevents the summary from adopting a partisan emotional fingerprint, it does not explicitly teach the model what a neutral fingerprint looks like. The synthetic summary should also adopt bias-free lexicons and writing styles inspired by expert-written summaries, which are generally regarded as less biased compared to news coverage produced by mass media. To achieve this, it is essential to distinguish the distinct writing styles of polarised news and the expert-written summaries. Within the semantic space, the model’s latent representations should be further optimised to align closely with the semantic embeddings of expert-written summaries while simultaneously distancing themselves from the embeddings of polarised news content.

Inspired by contrastive learning~\cite{Chen:20, Yan:21, Hjelm:18}, we utilize normalised temperature-scaled cross-entropy loss (NT-Xent)~\cite{Chen:20} as the contrastive loss and consider generated summaries as anchors. Expert-written summaries act as positive samples, pulling closer to anchors. While those news articles in the auxiliary dataset with strong political polarisation act as negative samples,  moving away from the anchors. The contrastive loss could be described as follows:

\begin{footnotesize}
\begin{align}
    \bar{H}^{D}_{Y} &= Avg(\text{Enc}_{\text{summ}}(Y^{D})) \\
    L_{Con} &=-log\frac{exp(sim(\bar{H}^{D}_{\text{summ}}, \bar{H}^{D}_{Y})/\tau)}{\sum_{H \in \{\bar{H}_{L}^{D_{\text{aux}}},\bar{H}_{R}^{D_{\text{aux}}},\bar{H}_{Y}^{D}\}}exp[sim(\bar{H}^{D}_{\text{summ}}, H)/\tau]}
\end{align}
\end{footnotesize}

Where $ \bar{H}^{D}_{Y}$ is the averaged encoded hidden state for the expert-written summary. $sim(\cdot, \cdot)$ denotes the similarity function and $\tau$ denotes the temperature. In the contrastive loss, the encoded embedding of summarised input $\bar{H}^{D}_{\text{summ}}$ and expert-written summary $\bar{H}^{D}_{Y}$ should be pulled closer to the semantic space. In contrast, the encoded summarised embedding should maintain a greater distance from the left and right political-leaning embeddings in the auxiliary dataset $\left\{ \bar{H}_{L}^{D_{\text{aux}}}, \bar{H}_{R}^{D_{\text{aux}}} \right\}$.

\subsection{Overall Model}
The two proposed losses are included in the fine-tuning process, where the BART pre-trained model is the backbone architecture. The overall loss function can be formulated as:


\begin{align}
    L_{Overall} = \lambda_{1} L_{MDS}+\lambda_{2} L_{ED}+\lambda_{3} L_{Con}
\end{align}

where $\lambda_{i}$ is the weight of each loss function, the ratio is $\frac{1}{3}:\frac{1}{3}:\frac{1}{3}$. The overall loss $L_{overall}$ consists of $L_{MDS}$, $L_{ED}$, and $L_{Con}$ representing multi-document summarisation loss, equal-distance loss, and contrastive loss, respectively.

\section{Media Bias Metric}
To evaluate NeutraSum's performance, we assess bias from two perspectives: the model's intrinsic ideological leaning and the emotional characteristics of its generated summaries. For content preservation, we adopt standard ROUGE and BLEU metrics.

\subsection{Political Compass Test Metric}
To gauge the inherent ideological leaning of the fine-tuned model itself, we adopted the Political Compass Test~\cite{Feng:23}. The underlying assumption is that a model with a strong intrinsic ideological bias is more likely to inject that bias, however subtly, into its generative outputs. However, we must explicitly state that this test evaluates the model's general disposition in response to a fixed set of political propositions, rather than the bias present in any single, topic-specific summary. Therefore, we employ this metric as a macro-level, auxiliary diagnostic tool. It complements our more granular, summary-specific metric by providing a broader characterisation of the model's baseline political orientation. Rooted in political spectrum theories~\cite{Eysenck:57, Rokeach:73}, this test maps the model’s responses to political, social, and economic statements onto a two-dimensional grid, capturing its orientation on both the economic (left-right) and social (authoritarian-libertarian) axes.

\subsection{Chain-of-thought LLM-based Metric}
Traditional text-based bias metrics~\cite{Trhlik:24, Liu:22b, ye:24, Bang:24} often lack explainability, failing to pinpoint the specific linguistic features contributing to bias. To overcome this, we propose a novel, explainable metric that quantifies the emotional fingerprint of a generated summary. This is achieved through a Chain-of-Thought (CoT)~\cite{Wei:22} process that leverages a Large Language Model (LLM) to deconstruct the summary's affective stance. The process systematically identifies the summary's topic, attitude, and the key vocabularies reflecting this stance.

\begin{tcolorbox}[colback=gray!10, boxrule=0.5pt, arc=0pt, outer arc=0pt, boxsep=2pt, left=2pt, right=2pt, top=2pt, bottom=2pt]
\textbf{Step 1 (Framing)} [\textit{Given same-story media reports} $X^{D}_{L}, X^{D}_{C}, X^{D}_{R}$, \textit{generated summary} $\hat{Y}^{D}$], what is the main character or topic in the generated summary $\hat{Y}^{D}$? What is the attitude towards it?
\end{tcolorbox}
After identifying the main character or topic $t$ and the corresponding attitude 
$a$ towards $t$, which can be supportive, denying, or neutral, we can iteratively explore the reasons $R$ behind this attitude:
\begin{tcolorbox}[colback=gray!10, boxrule=0.5pt, arc=0pt, outer arc=0pt, boxsep=2pt, left=2pt, right=2pt, top=2pt, bottom=2pt]
\textbf{Step 2 (Justification)} What are the reasons for attitude $a$ towards $t$ in the summary  $\hat{Y}^{D}$?
\end{tcolorbox}
With obtaining all the implicit understanding of the generated summary, we would extract the related words that indicate the attitude $a$ towards the main character or topic $t$ in the summary $\hat{Y}^{D}$:
\begin{tcolorbox}[colback=gray!10, boxrule=0.5pt, arc=0pt, outer arc=0pt, boxsep=2pt, left=2pt, right=2pt, top=2pt, bottom=2pt]
\textbf{Step 3 (Stance)} What are the related vocabularies that reflect the attitude $a$ towards the main character or topic $t$ in the summary $\hat{Y}^{D}$?
\end{tcolorbox}
Finally, based on the above answers, we ask the political leaning of the generated summary $\hat{Y}^{D}$:
\begin{tcolorbox}[colback=gray!10, boxrule=0.5pt, arc=0pt, outer arc=0pt, boxsep=2pt, left=2pt, right=2pt, top=2pt, bottom=2pt]
\textbf{Step 4 (Bias)} What is the political leaning of the summary $\hat{Y}^{D}$?
\end{tcolorbox}

The vocabularies extracted in Step 3 constitute the raw data for our emotional fingerprint analysis. To quantify the fingerprint, we score each word using the VAD emotional rating corpus~\cite{Moham:18, Lee:22} across its three core dimensions: \textit{valence} (emotional tone), \textit{arousal} (intensity), and \textit{dominance} (control). To capture the overall polarity and strategy, we isolate words with positive valence ($v > 0.65$) and negative valence ($v < 0.35$) to exclude those with neutral sentiment. Subsequently, we separately summed the selected positive and negative words to analyse their respective sentiment polarities. For extracted phrases, we would identify their synonyms and assign sentiment scores to these synonyms.

This approach, shown in Fig. \ref{fig:cot-framework}, aids in comprehending how various emotions are experienced and expressed, thereby offering a thorough explanation for the observed attitudes and political biases.

\begin{table}[t]
    \centering
    \begin{tabular}{|c|c|c|c|c|}
        \hline
                & Left  &  Right & centre & Total\\
       \hline
        Train& 3160  & 3160   & 3160   & 9480 \\
       \hline
        Val  &  395 & 395  & 395 & 1185  \\
       \hline
        Test     & 396 & 396   & 396   & 1188 \\
       \hline
       Total    & 3951  & 3951   & 3951   & 11853\\
       \hline
    \end{tabular}
    \caption{Allsides data statistics. Each triplet consists of a set of left, centre, and right news articles on the same event.}
    \label{table: NEUS dataset}
\end{table}


\subsection{Salient Information Preservation Metric}
To assess the performance of the generated summaries retrieving essential information from the input articles, we use the $\textrm{ROUGE}$~\cite{Lin:04} and $\textrm{BLEU}$~\cite{Papineni:02} to calculate the information preservation between generated summaries and expert-written summaries. $\textrm{ROUGE}$ is to measure how often n-grams in the machine-generated summary capture from the human-written summary, and $\textrm{BLEU}$ is to measure how often n-grams in the human summary appear in the human-written summary. It is noted that higher $\textrm{ROUGE}$ and $\textrm{BLEU}$ represent a better information preservation performance.

\section{Experiment}
\subsection{Dataset}
We leveraged data from Allsides Dataset~\cite{Lee:22} for the summarisation task. This dataset consists of a wide range of U.S political topics such as `Election', `Immigration', `Healthcare', `Abortion' , etc. The dataset incorporates 3464 triplets of news articles. Each triplet consists of left, centre, and right-wing news articles reporting the same issue, as well as an expert-written summary. It is mentioned that the ``centre''  still contains media bias, which does not represent bias-free~\cite{Allsides:04}. The dataset details are illustrated in Table \ref{table: NEUS dataset}. We randomly split this dataset into training, validation and testing. To integrate neutrality losses for the mitigation of media bias, we obtained the auxiliary dataset~\cite{Baly:20} that incorporates left-wing and right-wing news articles, covering many topics and media sources from Allsides. It is noted that the incorporation of an auxiliary dataset aims to guide the generated summary away from polarised texts while compacting the same-story news articles from diverse political perspectives. Allsides dataset and auxiliary dataset use MIT and Apache 2.0 license respectively.



\subsection{Parameter Setting}
The hyperparameters of the NeutraSum model during training are illustrated below: models for summarisation and neutralisation are Encoder-decoder and Encoder-only part of ``Bart-large'' in HuggingFace. Their corresponding datasets are Multi-document Dataset and Auxiliary Dataset. Multi-document Dataset contains a triplet of the left, centre, and right-wing news articles for summarisation tasks. The auxiliary dataset consists of different left and right-wing news articles to join neutrality losses. Learning rates for them are $\textit{3e-5}$ and $\textit{2e-5}$. As for the batch size, a sample from a Multi-document Dataset will be paired with two left and two right samples originating from the auxiliary dataset. The generated summarisation length is between 100 to 250. We ran all of the experiments on one NVIDIA 4090 GPU. As for the Chain-of-thought LLM-based Metric, we utilize the Llama 3.1-8B-Instruct to implement the chain-of-thought pipeline.

\subsection{Baselines}
Since we followed a summarisation task to do the neutral summary, we would conduct some experiments on some current state-of-the-art multi-document summarisation models:
\begin{enumerate}
    \item \textbf{LEXRANK}~\cite{Erkan:04}: An extractive single-document summarisation model that extracts the sentences with high similarity in documents, which is finetuned in DOC 2004~\cite{Over:02}.
    \item \textbf{BARTCNN}: An abstract single-document summarisation utilizing CNN/Dailymail~\cite{Hermann:15} to finetune on the BART-large~\cite{Lewis:19} model.
    \item \textbf{BARTMULTI}: A multi-document abstractive summarisation model, which is finetuned on the BART-large~\cite{Lewis:19} model using the Multi-News dataset~\cite{Fabbri:19}.
    \item \textbf{PEGASUSMULTI}: A multi-document summarisation model finetuned on pegasus model~\cite{Zhang:20} using Multi-News dataset.
    \item \textbf{NEUSFT}~\cite{Lee:21}: An abstract multi-document summarisation model that proposes this task to finetune the Allsides dataset.
    \item \textbf{NEUS-TITLE}~\cite{Lee:22}: An improved version of NEUSFT that adds title summarisation.
\end{enumerate}

\begin{table}[t]
\resizebox{0.5\textwidth}{!}{
\begin{tabular}{|c|cccc|}
\hline
\multirow{3}{*}{Model}  & \multicolumn{4}{c|}{Salient Information Preservation ($\uparrow$)}                                                  \\ \cline{2-5} 
                        & \multirow{2}{*}{BLEU} & \multirow{2}{*}{ROUGE1-R} & \multirow{2}{*}{ROUGE2-R}      & \multirow{2}{*}{ROUGEL-R}      \\
                        &                       &                           &                                &                                \\ \hline
All Source input        & 8.23                  & 56.68$\%$                 & 32.21$\%$                      & 54.17$\%$                      \\
LEXRANK                 & 12.21                 & 39.08$\%$                 & 17.66$\%$                      & 34.69$\%$                      \\
BARTCNN                 & 10.49                 & 35.63$\%$                 & 15.32$\%$                      & 32.22$\%$                      \\
PEGASUSMULTI            & 6.12                  & 44.42$\%$                 & 16.99$\%$                      & 39.45$\%$                      \\
BARTMULTI               & 4.24                  & 35.76$\%$                 & 12.48$\%$                      & 32.08$\%$                      \\
NEUSFT                  & 11.67                 & 35.11$\%$                 & 15.74$\%$                      & 31.43$\%$                      \\
NEUS-TITLE              & 12.05                 & 35.11$\%$                 & 16.47$\%$                      & 32.63$\%$                      \\ \hline
NeutraSum               & 11.99                 & \textbf{52.91\%}          & \uline{27.87$\%$} & \uline{48.08$\%$} \\
w/o Contrastive Loss    & \textbf{12.01}        & 52.28$\%$                 & \textbf{28.43$\%$}             & 48.14$\%$                      \\
w/o Equal-distance Loss & 11.71                 & \uline{52.87$\%$}                 & 28.33$\%$                      & \textbf{48.42$\%$}             \\ \hline
\end{tabular}}
 \caption{Experiment results for the baselines and the proposed models on the Allsides dataset. For Salient Information Preservation, higher BLEU and ROUGE-R ($\uparrow$) scores represent a better summarisation performance. Our hybrid NeutraSum and its variants (w/o Contrastive Loss and w/o Equal-distance Loss) show competitive good summarisation results. The best performance for each metric is highlighted in \textbf{bold}, while the second is \uline{underlined}.}
\label{table:NeutraSum}
\end{table}

\begin{table*}[t!]
    \centering
    \resizebox{0.70\textwidth}{!}{ 
    \begin{tabular}{|c|c|ccccc|}
    \hline 
    \multirow{2}{*}{Model}
      & Media Bias Metric ($\downarrow$) & \multicolumn{5}{c|}{Chain-of-thought LLM-based Metric ($\downarrow$)}  
      \\ \cline{2-7}
      
     ~ & \begin{tabular}[c]{@{}c@{}}Political Compass Test\\  (economic-axis, social-axis)\end{tabular} & V\_SCORE 
     & V\_POSITIVE &  V\_NEGATIVE & A\_SCORE & D\_SCORE  \\ \hline
        All Source input & \textbackslash{}  & 3.78 & 0.83 & 0.67 & 2.44 & 2.58 \\
        Expert-written Summary & \textbackslash{}  & 2.34 & 0.51 & 0.40 & 1.48 & 1.58 \\
        \hline
         LEXRANK & \textbackslash{}  & \uline{2.92} & 0.68 & 0.56 & 1.99 & 2.02 \\
        BARTCNN & (-0.38, 3.13) & 3.53 & 0.65 & 0.79 & 1.98 & 2.07 \\
        PEGASUSMULTI & (-1.3, 4.36) & 2.96 & \uline{0.61} & 0.48 & 1.75 & 1.97 \\ 
        BARTMULTI & (-1.25, 2.6) & 3.13 & 0.69 & 0.54 & 2.03 & 2.18 \\ 
        NEUSFT & (-1.25, 2.72) & 2.89 & 0.64 & 0.48 & 1.82 & 2.03 \\ 
        NEUS-TITLE & (-0.38, 3.13) & 2.90 & 0.65 & 0.46 & 1.81 & 2.01 \\ \hline
        NeutraSum & (\textbf{-0.38, 2.41}) & \textbf{2.88} & 0.65 & \textbf{0.42} & \textbf{1.72} & \textbf{1.95}\\ 
        w/o contrastive loss & (0.42, 2.61) & \uline{2.92} & \textbf{0.59} & \uline{0.44} & \uline{1.74} & \uline{1.96} \\ 
        w/o Equal-distance loss & (\textbf{0.38}, 2.51) & 3.08 & 0.68 & 0.45 & 1.77 & 2.06 \\ \hline
    \end{tabular}}
    \caption{NeutraSum (Allsides-tuned) attains the lowest in Political Compass Test with (-0.38, 2.41) and outperforms baselines in LLM-based metrics (v-score: 2.88 $\downarrow$, a/d-score: 1.72/1.95 $\downarrow$), demonstrating superior neutrality in summaries compared to other models. The coordinate diagram is shown in Fig.~\ref{fig:political-compass-test}. The best performance for each metric is highlighted in \textbf{bold}, while the second-best is \uline{underlined}.}
    \label{table:bias}
\end{table*}

\section{Result and Analysis}

In this section, we illustrated the experimental results of baselines and our model NeutraSum, which is shown in Table \ref{table:NeutraSum}. Based on the experimental performance in salient information preservation and media bias reduction, we pointed out the significant observations by quantitative analysis.

\subsection{Effectivness of neutrality losses in media bias mitigation and salient information preservation} 

As shown in Table \ref{table:NeutraSum}, NeutraSum outperforms other baselines in preserving salient information, with strong scores in BLEU (11.99) and competitive performance across ROUGE metrics, including ROUGE1-R (52.91\%) and ROUGEL-R (48.08\%). While variants of the model exhibit minor fluctuations in metrics—such as a slight BLEU increase to 12.01 or ROUGE-L improvement to 48.42\%—the core NeutraSum framework consistently balances high-quality summarisation with reduced bias.

Table \ref{table:bias} presents both the model bias scores and the media bias scores of the generated summaries produced by different models, evaluated using the Political Compass Test and Chain-of-Thought LLM-based Metric, which assess media bias based on the model's responses and sentiment based on the target entity of different issues. The coordinate in Political Compass Test represents the level of polarity in economic-axis and social-axis; the \textit{V\_SCORE}, \textit{A\_SCORE}, and \textit{D\_SCORE} represent the valence, arousal, and dominance scores measured by VAD rating corpus, respectively. Firstly, all source inputs consist of media reports on the same issue. The sentiment scores of media news are notably high, suggesting that news outlets often embed strong political emotions and biases in their reporting. In contrast, the expert-written summary exhibits the lowest sentiment scores, as it is crafted by media experts and is considered a bias-free target that serves as the ideal reference for learning. Secondly, compared with baseline models, it is observed that our proposed model, NeutraSum, achieves the lowest scores, with \textit{V\_SCORE} (2.88), \textit{A\_SCORE} (1.72), and \textit{D\_SCORE} (1.95), respectively. This indicates that the summaries generated by our model demonstrate a reduced reliance on emotional language and express minimal subjective opinions during framing, thereby maintaining a neutral stance toward specific topics or individuals. \textit{V\_POSITIVE} and  \textit{V\_NEGATIVE} illustrate the scores for the positive words ($v>0.65$) and negative words ($v < 0.35$). NeutraSum and its variants also achieve the lowest scores, with  \textit{V\_POSITIVE} of 0.59 and  \textit{V\_NEGATIVE} of 0.42. These results indicate that the summaries generated by NeutraSum exhibit a relatively moderate emotional polarity. Such performance is attributed to the incorporation of additional loss functions, which align the generated summaries with human-written summaries and separate their semantic space from those associated with left-wing and right-wing biases.

\begin{figure}[t]
\hspace{-1.5cm}
\centering
\includegraphics[scale=0.35]{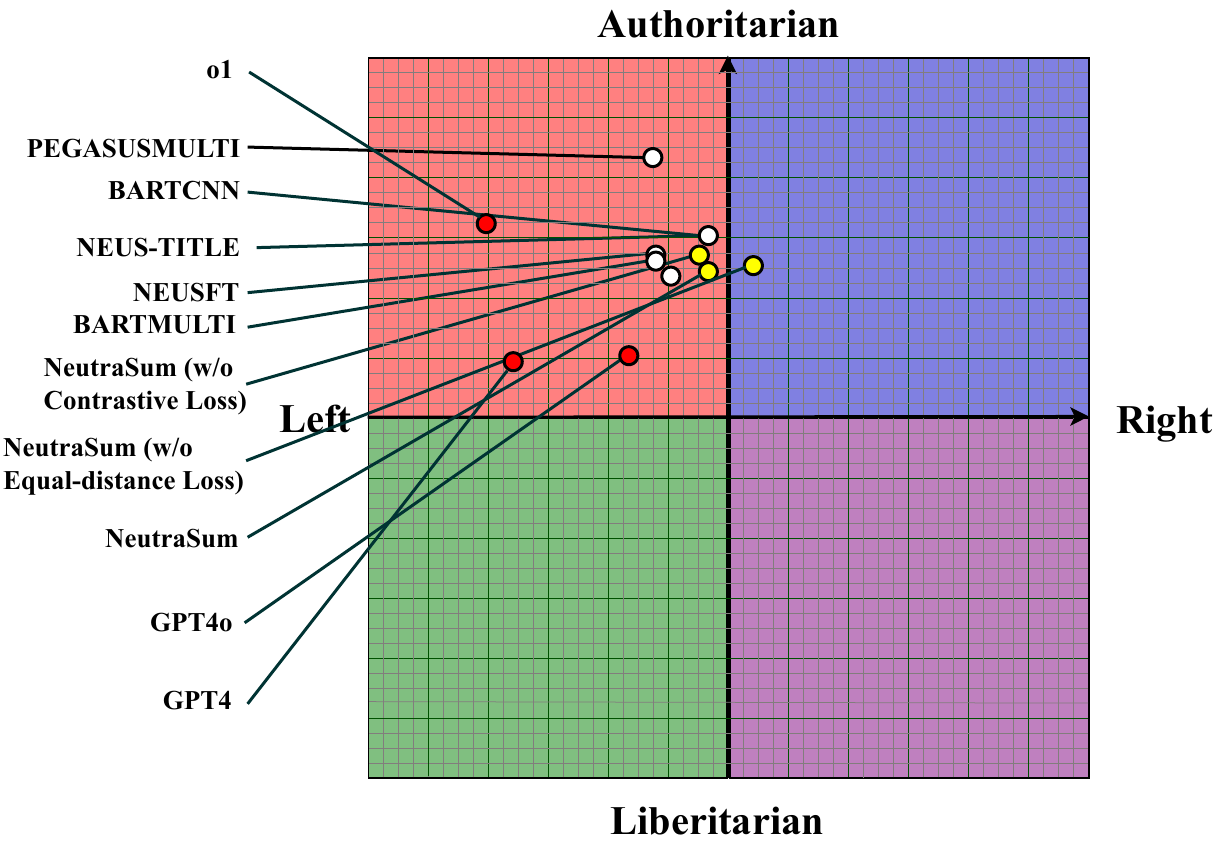}
\caption{Scores of different models in political compass test. The horizontal axis represents the economic axis (ranging from left to right), and the vertical axis represents the social axis (ranging from liberal to conservative). It could be seen that the yellow points of our model NeutraSum and its variants have relatively lower bias scores in both axes. We also take the bias score of GPT4/4o and o1 for reference.}
\label{fig:political-compass-test}
\end{figure}

\subsection{Necessity of two Neutrality Losses} 

The ablation study of NeutraSum demonstrates that while its variants achieve comparable summarisation quality, their bias mitigation capabilities differ markedly. Models trained without one of the neutrality losses exhibit instability in political bias reduction, with ambiguous responses in the political compass test causing economic-axis scores to drift moderately to $0.38$ or $0.42$. In contrast, the full NeutraSum model—equipped with both neutrality losses—reliably reduces media bias to a near-neutral score of $(-0.38, 2.41)$, underscoring the necessity of integrating both loss components for consistent and effective debiasing.


\subsection{Left and Authoritarian of the baselines in Polarity} 

From Fig. \ref{fig:political-compass-test}, it is observed that most of the baselines and our proposed models are slanted to the left and authoritarian in polarity based on the results of the political compass test. A contributing factor is that many media datasets crawled lots of news coverage from left-wing media outlets, such as CNN, the Guardian, BuzzFeed News, etc. summarisation models are fine-tuned on these news datasets and learn polarised writing styles. This proficiency is useful for detecting the political orientations of media content. However, this kind of feature adoption could potentially exacerbate the skew towards more leftist and authoritarian biases in media representation. Interestingly, we also test the popular large language models shown in red points. While GPT4, GPT4o and o1 specific models tend to lean towards left-authoritarian positions, GPT4 and GPT4o show comparatively moderate biases when addressing certain political issues. Conversely, o1 exhibits a distinctively stronger authoritarian bias, advocating for more comprehensive government oversight in both economic and social domains.

\begin{figure}[t]
    \centering
    \includegraphics[scale=0.5]{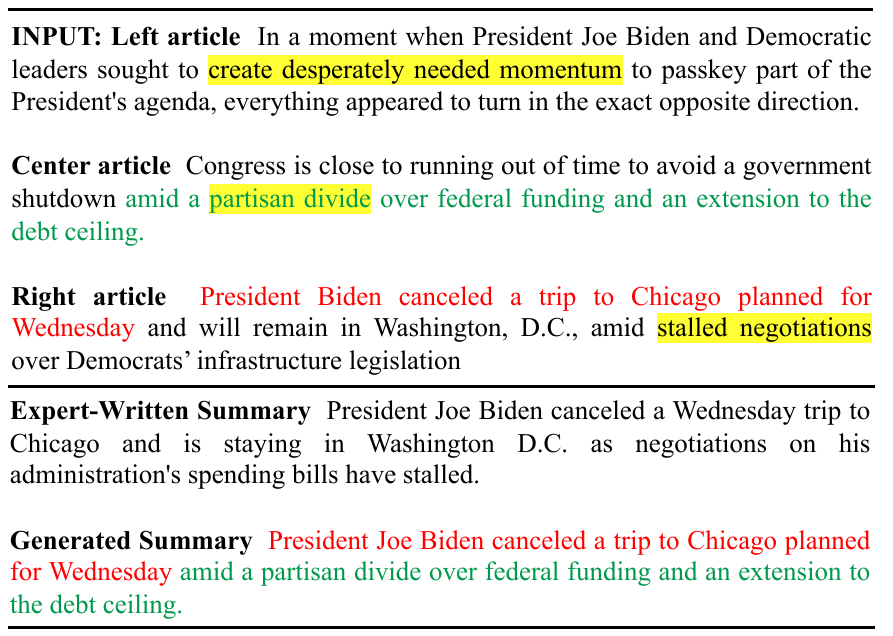}
    \caption{Illustration of generated summary, the corresponding input, and expert-written summary~\cite{Allsides:03}. \colorbox{yellow}{Yellow spans} illustrates the different descriptions of the attitudes and reasons for cancelling the president's agenda. The generated summary for this issue extracts the salient and objective sentences from the centre and right articles, which are highlighted in \textcolor{green}{green} and \textcolor{red}{red}.}
    \label{fig:gen-case-1}
\end{figure}

\begin{figure}[t!]
    \centering
    \includegraphics[scale=0.15]{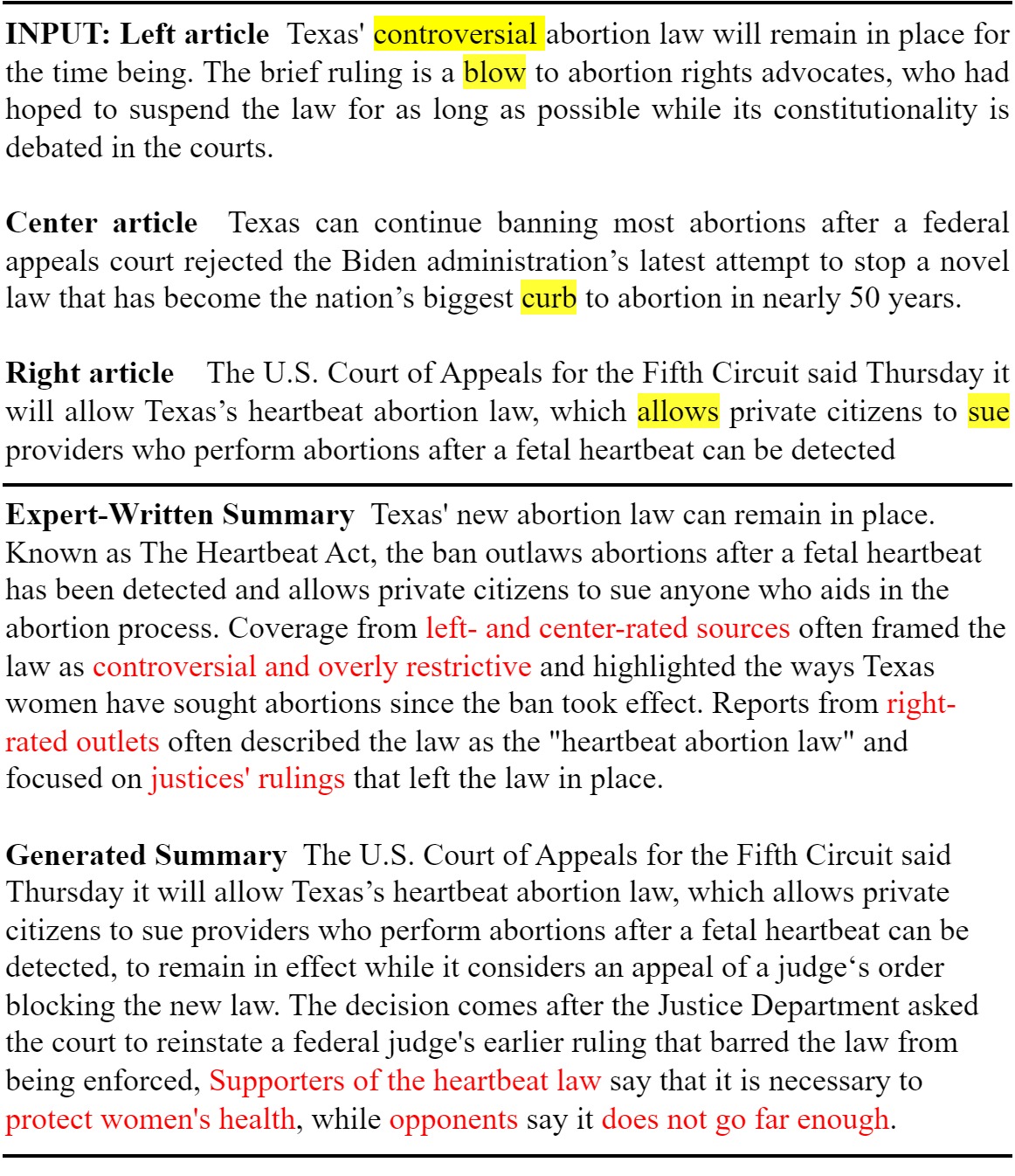}
    \caption{Illustration of generated summary, the corresponding input, and expert-written summary~\cite{Allsides:05}. \colorbox{yellow}{Yellow span} indicates that articles from various parties employ different words to articulate their positions on abortion laws. The generated summary not only outlines the full context of the issues but also encapsulates the perspectives and comments of various parties in \textcolor{red}{red}, similar to the expert-written summary.}
    \label{fig:gen-case-2}
\end{figure}

\subsection{Showcase of News Summarisation and Analysis}

As illustrated in Fig. \ref{fig:gen-case-1} and \ref{fig:gen-case-2}, we showed two cases of reporting of the same event, with input (left, centre, right news article), expert-written summary, and the generated summary from the NeutraSum model. By comparing these summaries and the source input, we demonstrate how the metric can serve as a powerful tool for analysing media bias.

In Fig. \ref{fig:gen-case-1}, all of the relevant news articles and summaries wrote about the issue of cancelation of the president's agenda, the left-wing article speculated that President Biden ``create desperately needed momentum'' to his agenda,  carries moderate-to-high emotional intensity due to words like ``desperately"(V=0.083, A=0.84, D=0.34) and ``momentum" (V=0.66, A=0.75, D=0.69), but ultimately shifts toward negativity as it highlights unexpected failure. While right-wing article stated that the reason for cancelling agenda is ``stalled negotiations over Democrats’ infrastructure legislation'', the low dominance of ''stalled" (V=0.37, A=0.25, D=0.29) implies that Biden or the Democrats lack full control over the situation, reinforcing the image of political gridlock.

An interesting pattern is shown in the generated summary. While BART is an automatic abstract summary model that will not directly copy the sentences from the input, we could observe that the generated summary extracts some bias-free sentences from the inputs to describe the issue. This seems to demonstrate the model's certain capacity and semantic understanding of media bias neutralisation.

Fig. \ref{fig:gen-case-2} still illustrates subtle word choice in different polarised articles and abstractive summarisation capacity. Regarding the coverage of abortion law issues, articles from the left-wing use the word ``controversial''(V=0.27, A=0.89, D=0.54) and ``blow''(V=0.32, A=0.75, D=0.57) to express the deny attitude toward the abortion law. These two words are both emotionally intense (high arousal) and slightly negative (low valence), which emphasize their negative impact on proponents of abortion rights. Articles from the centre share a comparable viewpoint with the left-wing, employing ``curb''(V=0.47, A=0.31, D=0.41) to depict the challenges associated with prohibiting abortion law. Given its neutral-to-slightly-negative valence and low arousal, the word choice presents the law as a significant limitation without overt emotional bias. Conversely, right-wing articles use terms like ``allows" (V=0.70, A=0.43, D=0.54) and ``sue" (V=0.22, A=0.73, D=0.68), reflecting a more favorable stance toward abortion legislation. The use of ``allows" suggests a positive or permissive framing of the law, while ``sue" carries high arousal and dominance, emphasizing the enforcement mechanism that empowers private citizens to take legal action. An interesting notion is that the generated summary, in its conclusion, presents a neutral perspective on the attitudes of both the left and right toward abortion laws. It does more than just summarise the event's overview, it also encapsulates the stances of all parties involved in the matter. More specifically, viewpoints of supporters use ``necessary'' and ``protest'' to emphasize the law as a moral imperative, portraying it as an action of care and responsibility rather than restriction. While  opponents' argument is framed with ``does not go far enough", a phrase synonymous with "inadequate" (V=0.12, A=0.45, D=0.23), implying dissatisfaction without proposing an alternative or stronger stance.

\section{Discussion}
This study pioneers a new framework for mitigating media bias by tracing partisan stances to their quantifiable ‘emotional fingerprints’. By reframing bias as a problem of neutralising specific VAD-based emotional patterns, our work moves beyond abstract political labels. Our proposed NeutraSum model, guided by novel neutrality losses, validates this approach. It successfully neutralises the identified emotional fingerprints—particularly the ``Forceful Critique" framework prominent in Left-leaning media—and achieves a demonstrably lower emotional bias score, thereby contributing to a more balanced media landscape.

Furthermore, our VAD-based metric, which leverages a Chain-of-Thought process, offers a significant advancement in explainable bias detection. Unlike opaque, probability-based methods, our metric directly identifies and scores the emotional language contributing to bias, providing a transparent tool for journalists and researchers. While the LLM-as-Judge may harbor intrinsic biases or exhibit instability, which can affect the reliability and robustness of the assessment, we frame it as a powerful explainable proxy. Its primary value lies not in producing a definitive score, but in its transparency. By surfacing the specific words and phrases that contribute to an emotional or partisan tone. This offers a diagnostic insight that opaque, probability-based metrics cannot provide, making it a valuable tool for understanding how bias manifests linguistically.

Nevertheless, our work has limitations. The reliance on the U.S.-centric Allsides dataset constrains cross-cultural generalizability, and our focus on semantic neutrality may overlook subtler tonal biases. Future work should expand this framework to multilingual, non-Western contexts and integrate more nuanced rhetorical analyses. Additionally, improving the computational efficiency of NeutraSum and enhancing its explainability will be crucial for its real-world application in combating societal polarisation.

\newpage 

\appendix

\noindent \textit{A. Analysis of the Weights for Losses.} The configuration of weights for different losses is a critical issue to discuss. We chose different weight combinations for the three losses and did the experiments in Table \ref{table: weight of losses}. From the first row of weight configuration, we adjust the weight of MDS Loss (Multi-document Summarisation Loss) progressively scaled down from 0.98 to 0.02. Meanwhile, the weights for neutrality losses, namely ED Loss (Equal-distance Loss) and Con Loss (Contrastive Loss), are weighted nearly identically due to their similar efficacy and significance.

The weight combination (0.2:0.5:0.3) in the NeutraSum yields relatively high summarisation results, while the (0.33:0.33:0.33) configuration effectively minimises media bias, achieving a score of (-0.38, 2.41) in the political compass test. Excessive weight for MDS Loss (0.98:0.01:0.01) might lead to more bias (-1.63, 3.96) in the summaries. In contrast, having a higher neutrality loss weight (0.02:0.49:0.49) does not necessarily ensure the best reduction in bias. It implies that there isn't a straightforward relationship where simply increasing the weight of these losses leads to less biased summaries. By considering the final effect observed in bias reduction, the weight configuration of (0.33:0.33:0.33) has been selected as the final setting. This is because it can reduce bias more effectively while maintaining a certain level of summarisation quality.

\begin{table}[htp!]
\resizebox{0.5\textwidth}{!}{
\begin{tabular}{|c|cccc|c|}
\hline
\multicolumn{1}{|c|}{\multirow{2}{*}{\begin{tabular}[c]{@{}c@{}}Weight of Loss in NeutraSum\\ ($\lambda_1:\lambda_2:\lambda_3$)\end{tabular}}} & \multicolumn{4}{c|}{Salient info ($\uparrow$)}                                                                      & framing bias metric ($\downarrow$)   \\ \cline{2-6}
\multicolumn{1}{|c|}{}                                                                                                                                         & \multicolumn{1}{l|}{BLEU}    & \multicolumn{1}{l|}{ROUGE1-R} & \multicolumn{1}{l|}{ROUGE2-R} & ROUGE-L & Political Compass Test \\ \hline
0.98:0.1:0.1                                                                                                                                                   & \multicolumn{1}{c|}{12.17} & \multicolumn{1}{c|}{53.02\%}  & \multicolumn{1}{c|}{28.02\%}  & 48.32\% & (-1.63, -3.95)           \\ \hline
0.7:0.2:0.1                                                                                                                                                    & \multicolumn{1}{c|}{12.20} & \multicolumn{1}{c|}{53.30\%}  & \multicolumn{1}{c|}{28.42\%}  & 48.60\% & (0.25, -4.0)              \\ \hline
0.33:0.33:0.33                                                                                                                                                 & \multicolumn{1}{c|}{11.99} & \multicolumn{1}{c|}{52.91\%}  & \multicolumn{1}{c|}{27.87\%}  & 48.08\% & \textbf{(-0.38, 2.41)}            \\ \hline
0.2:0.5:0.3                                                                                                                                                   & \multicolumn{1}{c|}{\textbf{12.24}} & \multicolumn{1}{c|}{\textbf{53.69\%}}  & \multicolumn{1}{c|}{\textbf{28.75\%}}  & \textbf{49.24\%} & (-0.63, -3.74)           \\ \hline
0.02:0.49:0.49                                                                                                                                                 & \multicolumn{1}{c|}{11.39} & \multicolumn{1}{c|}{52\%}     & \multicolumn{1}{c|}{27\%}     & 47.50\% & (0.50 ,-2.51)            \\ \hline
\end{tabular}}
\caption{Experimental results of adjusting different weights of losses (Multi-document Summarisation Loss:Equal-distance Loss:Contrastive Loss) in NeutraSum. The weight combination (0.2:0.5:0.4) illustrates the best summarisation performance in Salient Information Scores. However, weight combination (0.33:0.33:0.33) would be a better choice for weight configuration since it achieved the lowest bias score(-0.38, 2.41) in the political compass test and retained good summarisation results.}
\label{table: weight of losses}
\end{table}

\noindent \textit{B. Ethical Statement.} Media bias can be considered abstract in the sense that it involves subjective perceptions, opinions, and preferences within a political context. In this paper, we utilize the political compass test to measure media bias based on the general political background. However, this test could not assess the political attitude towards the specific event. While early research~\cite{Lee:22} has proposed sentiment-annotation lexicons to measure media bias by calculating the sentiment scores in the summary, there is no clear relation between sentiment and political polarity. Therefore, it is essential to be mindful of media bias considerations in both the model and the generated summaries in future research.

Another noteworthy phenomenon is that not all the same-story news articles with diverse political slants would have different attitudes. As for extreme cases, media outlets commonly adopt similar stances when reporting on particular issues. A broader investigative focus is warranted on relative polarity, which means that media outlets with various political ideologies have different attitudes toward the same issue. This could also be considered as a more nuanced perspective on future media bias metric design.

The final consideration is that to do this neutral summary task, we take out some notable cases to illustrate the framing of different polarised news outlets. It is noteworthy that not all news articles employ overly emphasized or derogatory language, this is primarily utilized to showcase the effectiveness of our summary generation.


\begin{thebibliography}{00}

\bibitem{eppler:15}
Eppler, M. 11. Information quality and information overload: The promises and perils of the information age. {\em Communication And Technology}. \textbf{5} pp. 215 (2015)

\bibitem{schmitt:18}
Schmitt, J., Debbelt, C. \& Schneider, F. Too much information? Predictors of information overload in the context of online news exposure. \emph{Information, Communication \& Society}. \textbf{21}, 1151-1167 (2018)


\bibitem{DA:20}
D.~D'Alessio and M.~Allen, ``Media bias in presidential elections: A meta-analysis,'' \emph{Journal of communication}, vol.~50, no.~4, pp. 133--156, 2000.


\bibitem{Rodrigo:24}
F.-J. Rodrigo-Gines, J.~Carrillo-de Albornoz, and L.~Plaza, ``A systematic review on media bias detection: What is media bias, how it is expressed, and how to detect it,'' \emph{Expert Systems with Applications}, vol. 237, p. 121641, 2024.

\bibitem{Gentzkow:06}
Matthew Gentzkow and Jesse~M Shapiro.
\newblock Media bias and reputation.
\newblock {\em Journal of political Economy}, 114(2):280--316, 2006.

\bibitem{Sunstein:01}
Cass~R Sunstein.
\newblock {\em Echo chambers: Bush v. Gore, impeachment, and beyond}.
\newblock Princeton University Press Princeton, NJ, 2001.


\bibitem{van2021fighting}
Meer, T. \& Hameleers, M. Fighting biased news diets: Using news media literacy interventions to stimulate online cross-cutting media exposure patterns. {\em New Media \& Society}. \textbf{23}, 3156-3178 (2021)

\bibitem{muise2022quantifying}
Muise, D., Hosseinmardi, H., Howland, B., Mobius, M., Rothschild, D. \& Watts, D. Quantifying partisan news diets in Web and TV audiences. {\em Science Advances}. \textbf{8}, eabn0083 (2022)

\bibitem{angelucci2024media}
Angelucci, C., Cagé, J. \& Sinkinson, M. Media competition and news diets. {\em American Economic Journal: Microeconomics}. \textbf{16}, 62-102 (2024)


\bibitem{Park:11}
Souneil Park, Minsam Ko, Jungwoo Kim, Ho-Jin Choi, and Junehwa Song.
\newblock Newscube 2.0: an exploratory design of a social news website for media bias mitigation.
\newblock In {\em Workshop on Social Recommender Systems}, pages 1--5, 2011.

\bibitem{picard1997ective}
Picard, R. \& Others A ective Computing. (MIT press Cambridge Massachusetts,1997)

\bibitem{Lee:22}
Nayeon Lee, Yejin Bang, Tiezheng Yu, Andrea Madotto, and Pascale Fung.
\newblock Neus: Neutral multi-news summarisation for mitigating framing bias.
\newblock {\em Annual Conference of the North American Chapter of the Association for Computational Linguistics (NAACL)}, 2022.

\bibitem{Zhang:19}
Yifan Zhang, Giovanni Da~San Martino, Alberto Barr{\'o}n-Cedeno, Salvatore Romeo, Jisun An, Haewoon Kwak, Todor Staykovski, Israa Jaradat, Georgi Karadzhov, Ramy Baly, et~al.
\newblock Tanbih: Get to know what you are reading.
\newblock {\em arXiv preprint arXiv:1910.02028}, 2019.

\bibitem{Bang:23}
Yejin Bang, Nayeon Lee, and Pascale Fung.
\newblock Mitigating framing bias with polarity minimisation loss.
\newblock {\em arXiv preprint arXiv:2311.01817}, 2023.

\bibitem{Trhlik:24}
F.~Trhlik and P.~Stenetorp, ``Quantifying generative media bias with a corpus of real-world and generated news articles,'' \emph{arXiv preprint arXiv:2406.10773}, 2024.

\bibitem{Liu:22b}
Ruibo Liu, Chenyan Jia, Jason Wei, Guangxuan Xu, and Soroush Vosoughi.
\newblock Quantifying and alleviating political bias in language models.
\newblock {\em Artificial Intelligence}, 304:103654, 2022.

\bibitem{ye:24}
J.~Ye, Y.~Wang, Y.~Huang, D.~Chen, Q.~Zhang, N.~Moniz, T.~Gao, W.~Geyer, C.~Huang, P.-Y. Chen \emph{et~al.}, ``Justice or prejudice? quantifying biases in llm-as-a-judge,'' \emph{arXiv preprint arXiv:2410.02736}, 2024.

\bibitem{Elejalde:18}
E.~Elejalde, L.~Ferres, and E.~Herder, ``On the nature of real and perceived bias in the mainstream media,'' \emph{PloS one}, vol.~13, no.~3, p. e0193765, 2018.

\bibitem{Wei:22}
J.~Wei, X.~Wang, D.~Schuurmans, M.~Bosma, F.~Xia, E.~Chi, Q.~V. Le, D.~Zhou \emph{et~al.}, ``Chain-of-thought prompting elicits reasoning in large language models,'' \emph{Advances in neural information processing systems}, vol.~35, pp. 24\,824--24\,837, 2022.

\bibitem{Moham:18}
S.~Mohammad, ``Obtaining reliable human ratings of valence, arousal, and dominance for 20,000 english words,'' in \emph{Proceedings of the 56th annual meeting of the association for computational linguistics (volume 1: Long papers)}, 2018, pp. 174--184.

\bibitem{Hamborg:19}
Felix Hamborg, Karsten Donnay, and Bela Gipp.
\newblock Automated identification of media bias in news articles: an interdisciplinary literature review.
\newblock {\em International Journal on Digital Libraries}, 20(4):391--415, 2019.

\bibitem{Liu:22a}
Yujian Liu, Xinliang~Frederick Zhang, David Wegsman, Nick Beauchamp, and Lu~Wang.
\newblock Politics: pretraining with same-story article comparison for ideology prediction and stance detection.
\newblock {\em arXiv preprint arXiv:2205.00619}, 2022.

\bibitem{Baly:20}
Ramy Baly, Giovanni Da~San Martino, James Glass, and Preslav Nakov.
\newblock We can detect your bias: Predicting the political ideology of news articles.
\newblock {\em arXiv preprint arXiv:2010.05338}, 2020.

\bibitem{Fan:19}
Lisa Fan, Marshall White, Eva Sharma, Ruisi Su, Prafulla~Kumar Choubey, Ruihong Huang, and Lu~Wang.
\newblock In plain sight: Media bias through the lens of factual reporting.
\newblock {\em arXiv preprint arXiv:1909.02670}, 2019.

\bibitem{Hamborg:17}
Felix Hamborg, Norman Meuschke, and Bela Gipp.
\newblock Matrix-based news aggregation: exploring different news perspectives.
\newblock In {\em 2017 ACM/IEEE Joint Conference on Digital Libraries (JCDL)}, pages 1--10. IEEE, 2017.

\bibitem{Van:20}
Esther van~den Berg and Katja Markert.
\newblock Context in informational bias detection.
\newblock In {\em Proceedings of the 28th International Conference on Computational Linguistics}, pages 6315--6326, 2020.

\bibitem{Gangula:19}
Rama Rohit~Reddy Gangula, Suma~Reddy Duggenpudi, and Radhika Mamidi.
\newblock Detecting political bias in news articles using headline attention.
\newblock In {\em Proceedings of the 2019 ACL workshop BlackboxNLP: analyzing and interpreting neural networks for NLP}, pages 77--84, 2019.

\bibitem{peng2024beyond}
Peng, T., Yang, K., Lee, S., Li, H., Chu, Y., Lin, Y. \& Liu, H. Beyond Partisan Leaning: A Comparative Analysis of Political Bias in Large Language Models. {\em ArXiv Preprint ArXiv:2412.16746}. (2024)


\bibitem{Spinde:21}
Timo Spinde, Lada Rudnitckaia, Kanishka Sinha, Felix Hamborg, Bela Gipp, and Karsten Donnay.
\newblock Mbic--a media bias annotation dataset including annotator characteristics.
\newblock {\em arXiv preprint arXiv:2105.11910}, 2021.

\bibitem{Maab:24}
I.~Maab, E.~Marrese-Taylor, S.~Pad{\'o}, and Y.~Matsuo, ``Media bias detection across families of language models,'' in \emph{Proceedings of the 2024 Conference of the North American Chapter of the Association for Computational Linguistics: Human Language Technologies (Volume 1: Long Papers)}, 2024, pp. 4083--4098.

\bibitem{Lin:24}
L.~Lin, L.~Wang, X.~Zhao, J.~Li, and K.-F. Wong, ``Indivec: An exploration of leveraging large language models for media bias detection with fine-grained bias indicators,'' \emph{arXiv preprint arXiv:2402.00345}, 2024.

\bibitem{Chen:24}
L.~Chen and G.~Varoquaux, ``What is the role of small models in the llm era: A survey,'' \emph{arXiv preprint arXiv:2409.06857}, 2024.

\bibitem{Wang:24}
F.~Wang, Z.~Zhang, X.~Zhang, Z.~Wu, T.~Mo, Q.~Lu, W.~Wang, R.~Li, J.~Xu, X.~Tang \emph{et~al.}, ``A comprehensive survey of small language models in the era of large language models: Techniques, enhancements, applications, collaboration with llms, and trustworthiness,'' \emph{arXiv preprint arXiv:2411.03350}, 2024.

\bibitem{Park:09}
Souneil Park, Seungwoo Kang, Sangyoung Chung, and Junehwa Song.
\newblock Newscube: delivering multiple aspects of news to mitigate media bias.
\newblock In {\em Proceedings of the SIGCHI conference on human factors in computing systems}, pages 443--452, 2009.

\bibitem{Trampuvs:15}
Mitja Trampu{\v{s}}, Flavio Fuart, Daniele Pighin, Tadej {\v{S}}tajner, Jan Ber{\v{c}}i{\v{c}}, Blaz Novak, Delia Rusu, Luka Stopar, and Marko Grobelnik.
\newblock Diversinews: surfacing diversity in online news.
\newblock {\em AI Magazine}, 36(4):87--104, 2015.

\bibitem{Lee:21}
Nayeon Lee, Yejin Bang, Andrea Madotto, and Pascale Fung.
\newblock Mitigating media bias through neutral article generation.
\newblock {\em arXiv preprint arXiv:2104.00336}, 2021.

\bibitem{Liu:21a}
Ruibo Liu, Chenyan Jia, and Soroush Vosoughi.
\newblock A transformer-based framework for neutralising and reversing the political polarity of news articles.
\newblock {\em Proceedings of the ACM on Human-Computer Interaction}, 5(CSCW1):1--26, 2021.

\bibitem{Allsides:04}
Allsides.
\newblock What does a "centre" media bias rating mean?
\newblock Website, 2019.

\bibitem{Liu:21b}
Ruibo Liu, Lili Wang, Chenyan Jia, and Soroush Vosoughi.
\newblock Political depolarisation of news articles using attribute-aware word embeddings.
\newblock In {\em Proceedings of the International AAAI Conference on Web and Social Media}, volume~15, pages 385--396, 2021.

\bibitem{Chen:20}
Wei-Fan Chen, Khalid Al-Khatib, Benno Stein, and Henning Wachsmuth.
\newblock Detecting media bias in news articles using gaussian bias distributions.
\newblock {\em arXiv preprint arXiv:2010.10649}, 2020.

\bibitem{Yan:21}
Yuanmeng Yan, Rumei Li, Sirui Wang, Fuzheng Zhang, Wei Wu, and Weiran Xu.
\newblock Consert: A contrastive framework for self-supervised sentence representation transfer.
\newblock {\em arXiv preprint arXiv:2105.11741}, 2021.

\bibitem{Hjelm:18}
R~Devon Hjelm, Alex Fedorov, Samuel Lavoie-Marchildon, Karan Grewal, Phil Bachman, Adam Trischler, and Yoshua Bengio.
\newblock Learning deep representations by mutual information estimation and maximisation.
\newblock {\em arXiv preprint arXiv:1808.06670}, 2018.

\bibitem{Feng:23}
Shangbin Feng, Chan~Young Park, Yuhan Liu, and Yulia Tsvetkov.
\newblock From pretraining data to language models to downstream tasks: Tracking the trails of political biases leading to unfair nlp models.
\newblock {\em arXiv preprint arXiv:2305.08283}, 2023.

\bibitem{Eysenck:57}
Hans~Jurgen Eysenck.
\newblock Sense and nonsense in psychology.
\newblock {\em Penguin Books}, 1957.

\bibitem{Rokeach:73}
Milton Rokeach.
\newblock {\em The nature of human values.}
\newblock Free press, 1973.

\bibitem{Bang:24}
Y.~Bang, D.~Chen, N.~Lee, and P.~Fung, ``Measuring political bias in large language models: What is said and how it is said,'' \emph{arXiv preprint arXiv:2403.18932}, 2024.

\bibitem{Lin:04}
Chin-Yew Lin.
\newblock Rouge: A package for automatic evaluation of summaries.
\newblock In {\em Text summarisation branches out}, pages 74--81, 2004.

\bibitem{Papineni:02}
Kishore Papineni, Salim Roukos, Todd Ward, and Wei-Jing Zhu.
\newblock Bleu: a method for automatic evaluation of machine translation.
\newblock In {\em Proceedings of the 40th annual meeting of the Association for Computational Linguistics}, pages 311--318, 2002.

\bibitem{Erkan:04}
G{\"u}nes Erkan and Dragomir~R Radev.
\newblock Lexrank: Graph-based lexical centrality as salience in text summarisation.
\newblock {\em Journal of artificial intelligence research}, 22:457--479, 2004.

\bibitem{Over:02}
Paul Over and J~Yen.
\newblock An introduction to duc 2004 intrinsic evaluation of generic new text summarisation systems, 2004.
\newblock {\em National Institute of Standards and Technology}, 2002.

\bibitem{Hermann:15}
Karl~Moritz Hermann, Tomas Kocisky, Edward Grefenstette, Lasse Espeholt, Will Kay, Mustafa Suleyman, and Phil Blunsom.
\newblock Teaching machines to read and comprehend.
\newblock {\em Advances in neural information processing systems}, 28, 2015.

\bibitem{Lewis:19}
Mike Lewis, Yinhan Liu, Naman Goyal, Marjan Ghazvininejad, Abdelrahman Mohamed, Omer Levy, Ves Stoyanov, and Luke Zettlemoyer.
\newblock Bart: Denoising sequence-to-sequence pre-training for natural language generation, translation, and comprehension.
\newblock {\em arXiv preprint arXiv:1910.13461}, 2019.

\bibitem{Fabbri:19}
Alexander~R Fabbri, Irene Li, Tianwei She, Suyi Li, and Dragomir~R Radev.
\newblock Multi-news: A large-scale multi-document summarisation dataset and abstractive hierarchical model.
\newblock {\em arXiv preprint arXiv:1906.01749}, 2019.

\bibitem{Zhang:20}
Jingqing Zhang, Yao Zhao, Mohammad Saleh, and Peter Liu.
\newblock Pegasus: Pre-training with extracted gap-sentences for abstractive summarisation.
\newblock In {\em International Conference on Machine Learning}, pages 11328--11339. PMLR, 2020.

\bibitem{Allsides:05}
Allsides.
\newblock Federal appeals court keeps texas abortion ban intact; doj asks supreme court to halt law.
\newblock Website, 2021.


\bibitem{Allsides:03}
Allsides.
\newblock Biden cancels trip, focuses on spending bills as votes approach.
\newblock Website, 2021.
\newblock last accessed 21th September 2021.

\end{thebibliography}
\end{document}